\documentclass[letterpaper, 10 pt, conference]{ieeeconf}
\IEEEoverridecommandlockouts
\usepackage{graphics} % for pdf, bitmapped graphics files
\usepackage{epsfig} % for postscript graphics files
\usepackage{times} % assumes new font selection scheme installed
\usepackage{amsmath} % assumes amsmath package installed
\usepackage{amssymb}  % assumes amsmath package installed
\usepackage{url}
\usepackage{soul}
\usepackage{amsmath} 
\usepackage{subfigure}
\usepackage{multicol,tabularx,capt-of}
\usepackage{multirow}
\usepackage{mathtools}
\usepackage{color}

\title{\LARGE \bf
Learning from Demonstrations for Autonomous Soft-tissue Retraction*
}

\author{Ameya Pore$^{1,2}$, Eleonora Tagliabue$^{1}$, Marco Piccinelli$^{1}$, Diego Dall'Alba$^{1}$, Alicia Casals$^{2}$, Paolo Fiorini$^{1}$ % <-this % stops a space
\thanks{$^{1}$ Department of Computer Science, University of Verona, Italy. Email: {\tt\small ameya.pore@univr.it}}%
\thanks{$^{2}$ Automatic Control and Computer Engineering Department, Universitat Politècnica de Catalunya, Barcelona, Spain }%
\thanks{*This project has received funding from the European Union’s Horizon 2020 research and innovation programme under the Marie Skłodowska-Curie (grant agreement No. 813782 "ATLAS") and (grant agreement No. 742671 "ARS")}
}

\begin{document}

\maketitle
\thispagestyle{empty}
\pagestyle{empty}

% \pagestyle{plain}
% \setcounter{page}{1}
% \pagenumbering{arabic}

%%%%%%%%%%%%%%%%%%%%%%%%%%%%%%%%%%%%%%%%%%%%%%%%%%%%%%%%%%%%%%%%%%%%%%%%%%%%%%%%
\begin{abstract}

The current research focus in Robot-Assisted Minimally Invasive Surgery (RAMIS) is directed towards increasing the level of robot autonomy, to place surgeons in a supervisory position.
Although Learning from Demonstrations (LfD) approaches are among the preferred ways for an autonomous surgical system to learn expert gestures, they require a high number of demonstrations and show poor generalization to the variable conditions of the surgical environment. 
In this work, we propose an LfD methodology based on Generative Adversarial Imitation Learning (GAIL) that is built on a Deep Reinforcement Learning (DRL) setting. 
GAIL combines generative adversarial networks to learn the distribution of expert trajectories with a DRL setting to ensure generalisation of trajectories providing human-like behaviour.
% GAIL uses generative adversarial networks to learn the distribution of expert trajectories and ensures generalisation to diversified trajectories with human-like behaviour at the same time, thanks to the DRL setting.
We consider automation of tissue retraction, a common RAMIS task that involves soft tissues manipulation to expose a region of interest. 
In our proposed methodology, a small set of expert trajectories can be acquired through the da Vinci Research Kit (dVRK) and used to train the proposed LfD method inside a simulated environment. 
Results indicate that our methodology can accomplish the tissue retraction task with human-like behaviour while being more sample-efficient than the baseline DRL method. 
Towards the end, we show that the learnt policies can be successfully transferred to the real robotic platform and deployed for soft tissue retraction on a synthetic phantom.

\end{abstract}

%%%%%%%%%%%%%%%%%%%%%%%%%%%%%%%%%%%%%%%%%%%%%%%%%%%%%%%%%%%%%%%%%%%%%%%%%%%%%%%%
\section{INTRODUCTION\label{sec:intro}}
Robot-Assisted Minimally Invasive Surgery (RAMIS) is a consolidated paradigm in the field of medical robotics. RAMIS is a viable alternative to other surgical approaches that reduce several patient complications such as excessive intraoperative blood loss, post-operative trauma, and a high amount of mortality associated with more invasive surgeries \cite{simaan2018medical}. One of the widely adopted platforms for RAMIS is the da Vinci Surgical System (Intuitive Surgical Inc.), a teleoperated system that enables dexterous control with enhanced precision, stability and accuracy.

\begin{figure}[t]
	\centering
	\includegraphics[width=0.49\textwidth]{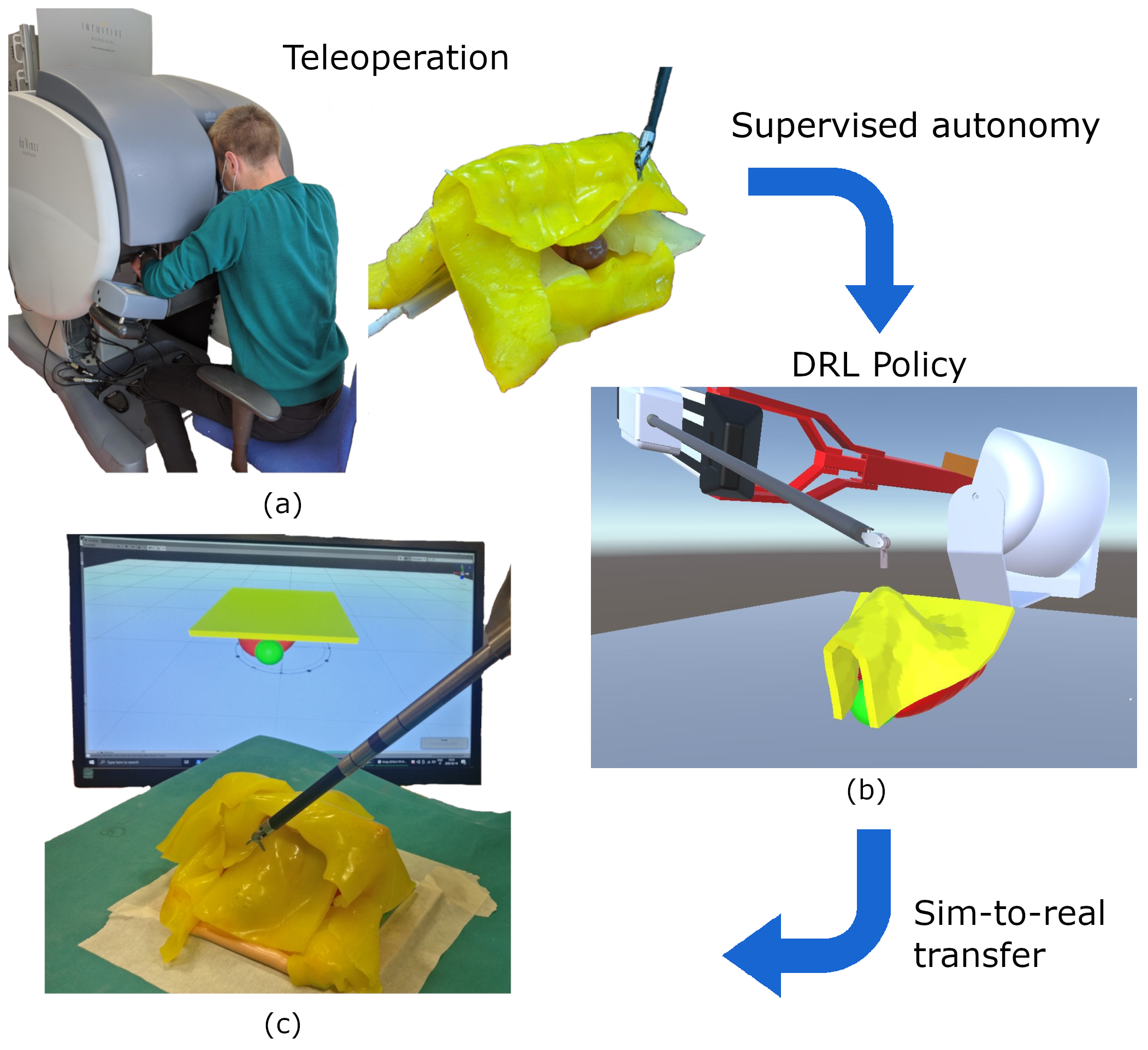}
	\caption{The proposed methodology of LfD for the tissue retraction surgical gesture. (a) Expert demonstrations are performed and recorded using the dVRK console %and recorded in \textit{UnityFlexML} using a communication pipeline. 
	(b) Robotic agent is trained within a simulated environment. (c) The learnt policy is translated to the real robotic system.}
	\label{fig:setup}
\end{figure}

A significant portion of RAMIS procedures is spent in mobilizing and manipulating tissues to reach the region of interest.
The tissue is repetitively grasped and retracted to expose the underlying anatomical area \cite{patil2010toward}. This gesture of tissue retraction occurs in multiple phases of surgery and
requires extensive interaction with soft tissues having heterogeneous physical and geometric properties, such as stiffness and viscoelasticity, with high inter and intra-subject variability. During RAMIS, tissue retraction task may require the surgeon either to switch robotic arms during surgery with a different set of visuomotor feedback and limited perception or to instruct an assistant with the desired motion \cite{attanasio2020autonomous}. This involves risks such as instrument collision or tissues damage that can negatively affect the procedure. Therefore, soft tissue retraction is an ideal candidate for the automation of surgical gestures since it would help to reduce the cognitive and physical workload of the surgeons and place them in a supervisory position.

%%%%%%% ELEONORA REPHRASING:
% A significant portion of RAMIS is spent in mobilization and manipulation of soft tissues to reach an underlying anatomical area, a task which is called tissue retraction \cite{patil2010toward}.
% This surgical task is particularly challenging because it requires extensive interaction with the anatomical tissues, characterized by heterogeneous geometric and physical properties, such as stiffness and viscoelasticity, with high inter- and intra-subject variability. During RAMIS, tissue retraction task requires the surgeon either to switch robotic arms with a different set of visuomotor feedback and limited perception or to instruct an assistant with the desired motion \cite{attanasio2020autonomous}, which might lead to instrument collision or tissues damage that can negatively affect the outcome of the procedure. Therefore, soft tissue retraction is an ideal candidate for the automation of surgical gestures, since it would help to reduce the cognitive and physical workload of the surgeons and place them in a supervisory position. 

% In the automation of surgical gestures, standard motion planning techniques, such as potential functions and splines tend to be successful in static conditions \cite{rimon1992exact, magid2006spline}. 
Standard motion planning techniques such as potential fields have been proposed for the automation of surgical gestures \cite{ginesi2019dynamic}. Such methods have shown to be successful in static conditions, but fail to generalize to dynamic environments like the surgical one, where human-level dexterity and adaptability are required.
% Standard motion planning techniques such as potential functions and splines tend to be successful in static conditions \cite{rimon1992exact, magid2006spline}. 
Hence, Learning from Demonstrations (LfD) is a preferred way to learn human gestures. 
% Several LfD methods have been developed using Gaussian mixture models \cite{reiley2010motion, su2021toward} and supervised learning methods such as behaviour cloning \cite{zhang2018deep}. %and Others \cite{argall2009survey}. 
However, a significant drawback of LfD methods is that they require a huge number of demonstrations to be trained properly, which is unfeasible in clinical settings considering the time, resources and ethical constraints.
% which is infeasible in the surgical paradigm given the safety and ethical concerns.
Moreover, LfD methods are affected by several limitations, such as errors in the acquisition and poor generalisation performance \cite{hussein2017imitation}.
% since these methods are purely based on LfD, they are affected by several limitations, such as errors in the acquisition and poor generalisation performance \cite{hussein2017imitation}. 
Using LfD, the robot can only become as good as the human's demonstrations. There is no additional information for improving the learnt behaviour. 

%%%%%%%%%
% A major limitation of LfD is that the robot can only become as good as the human's demonstrations. There is no additional information for improving the learnt behavior. Reinforcement learning, in contrast, allows the robot to discover new control policies through free exploration of the state-action space, but often takes a long time to converge. Approaches that combine the two aim at exploiting the strength of both to overcome their respective drawbacks. Particularly, demonstrations are used to initiate and guide the exploration done during reinforcement learning, reducing the time to find an improved control policy, which may depart from the demonstrated behavior.

By contrast, Deep Reinforcement Learning (DRL) allows the robot to discover new control policies through free exploration of the state-action space. DRL has shown generalisation capabilities in learning adaptable behaviours for complex and diverse scenarios such as dexterous manipulation and grasping \cite{ibarz2021train}. However, DRL methods often take a long time to converge and require a well-shaped and carefully designed reward function to learn simple goal-directed behaviours. Approaches that combine LfD and DRL aim at exploiting the strength of both to overcome their respective drawbacks. Particularly, demonstrations can be used to guide the exploration done during learning, reducing the time required to find an improved control policy, which may depart from the demonstrated behaviour.

% Deep Reinforcement Learning (DRL) has shown significant generalisation capabilities in learning adaptable behaviours for complex and diverse scenarios such as dexterous manipulation and grasping \cite{ibarz2021train}.
% % Deep Reinforcement Learning (DRL) has shown significant success in learning adaptable robotic solutions in complex and diverse scenarios such as dexterous manipulation and grasping \cite{ibarz2021train}. 
% However, a major drawback of DRL methods is the requirement of a well-shaped and carefully designed reward function while taking millions of trial and error attempts to learn simple goal-directed behaviours. This is due to the fact that DRL explores the behaviour space without any prior task-related knowledge \cite{dubey2018investigating}. Hence, demonstrations from expert users can be used to provide a preconceived reference such that the policy learnt can mimic human-like gestures while demonstrating adaptability. 

In this work, we address the limitations of LfD approaches by proposing a training methodology for surgical robots based on Generative Adversarial Imitation Learning (GAIL).
GAIL relies on generative adversarial networks to generate trajectories that are similar to the acquired expert demonstrations \cite{ho2016generative}.
% GAIL uses DRL to generate trajectories that are close to the acquired expert demonstrations using Generative adversarial Networks \cite{ho2016generative}.
% We target the automation of soft tissue retraction in the context of robot-assisted partial nephrectomy. During the procedure, the surgeon has to grasp a highly deformable perinephric fat tissue to expose an underlying tumour. 
To automate soft tissue retraction, we train GAIL in a simulated replica of the real environment using a small set of demonstrations acquired through the real robotic platform via teleoperation. This work represents an initial step for utilizing expert demonstrations to learn a control policy and to transfer the motion skills to a real robotic manipulator.
Hence, our contribution is the introduction of an LfD training methodology to learn human-level surgical gestures using a small set of task demonstrations.

In Sec.~\ref{sec:related_works}, we provide an overview of the related works. We summarise the mathematical formulation of the DRL methods used in Sec.~\ref{sec:background}. In Sec.~\ref{sec:methods}, we detail the training methodology and the learning setup. In Sec.~\ref{sec:experiments}, we describe experiments conducted to validate our methodology. Further, we present the results in Sec.~\ref{sec:results} along with discussion. In the final Sec.~\ref{sec:conclusion}, we elaborate the conclusions and highlight our future works. 

\section{RELATED WORKS\label{sec:related_works}}

Previous works to automate soft-tissue retraction use motion planning algorithms such as probabilistic roadmaps for optimization based objectives \cite{patil2010toward}. 
This method works well for pre-operative planning in a static environment. Nagy et al. developed an approach for tissue retraction based on images, where three methods based on proportional control, Hidden Markov Models and fuzzy logic are validated \cite{nagy2018surgical}. Further, Attanasio et al. developed a trajectory planner based on coordinates extracted from images \cite{attanasio2020autonomous}. Both of these studies require hand-crafting control strategies and movement sequences. The execution of complex non-linear trajectories and behaviours may be challenging using these methods.

\textit{DRL for surgical tasks}: Prior works have used DRL to learn the tensioning policy for tissue manipulation \cite{nguyen2019manipulating}. Pedram et al. developed a Q-learning RL algorithm based on visual features to learn a control policy for soft-tissue manipulation.
Shin et al. developed an adaptive Model Predictive Controller (MPC) to learn the tissue dynamics \cite{shin2019autonomous}. Human demonstrations are used to pretrain and initialize the tissue dynamics model. The learnt tensioning policy is used to develop a visual model-based RL algorithm that outputs actions to manipulate the tissue points to pre-specified locations. In the former work, the tissue dynamics is learnt implicitly in the DRL setting, whereas in the latter, the dynamics is learnt via an MPC. These methods would be well suited for learning dynamics of deformable objects with similar physical properties. However, variability in soft tissues properties would require complete re-training of these methods for each subject. Hence, we base our hypothesis on learning task features of tissue manipulation from expert demonstrations in contrast to learning tissue dynamics. Task features would inherently include the surgeon's knowledge in manipulating a variety of tissues and would demonstrate better generalisation.

In the context of evaluation platform, Richter et al. proposed a framework for training DRL policies in simulation for the dVRK system and showed that the learnt policy could be transferred to a real robotic system \cite{richter2019open}. The proposed environment does not support deformable objects, which is a major limitation when simulating the surgical scenario. Recently, Tagliabue et al. proposed a virtual framework called \textit{UnityFlexML} for simulating deformable tissues that is well suited to train DRL methods \cite{tagliabue2020soft}. We use this simulation framework to develop our LfD approach.  

\textit{LfD for surgical tasks}: 
Reiley et al. proposed a demonstration based framework using Gaussian Mixture Models (GMM) for motion generation \cite{reiley2010motion}. Recently, a similar approach of GMM has been used to learn dynamic motion primitives from the demonstrations obtained from expert surgeons \cite{su2021toward}. Osa et al. introduced an iterative technique to learn a single reference trajectory for knot tying \cite{osa2017online}. However, a single demonstration does provide enough consistency to model a manipulation skill. Schulman \textit{et al}. used a trajectory transfer algorithm to learn from demonstrations for the task of suturing \cite{schulman2013case}. Murali \textit{et al}. devised a method to segment demonstrations into motion sequences \cite{murali2015learning}. 
Standard approaches for imitation learning such as Behaviour Cloning (BC) and Inverse Reinforcement Learning (IRL), tend to be successful with large amounts of data. However, they suffer from compounding error caused by covariant shift and are extremely expensive to train \cite{ibarz2021train}, hence not suited for the surgical paradigm. 
An emerging derivative of IRL is represented by GAIL \cite{ho2016generative}. Contrary to IRL, which learns a cost function, GAIL methods learn how to act by directly learning the policy. GAIL has shown an empirical improvement in reducing the number of demonstrations required to successfully learn the task, compared to other imitation learning methods. 
Therefore we selected GAIL as the LfD method, since it has been successfully applied in other surgical domains (e.g., endovascular manipulators \cite{chi2020collaborative}) but never used in the RAMIS context.
% To the best of the authors' knowledge, GAIL has been proposed in the surgical field only for the control of endovascular manipulators \cite{chi2020collaborative}.

\section{BACKGROUND\label{sec:background}}

A general Reinforcement Learning (RL) problem can be formulated in terms of a Markov Decision Process (MDP), where an agent learns by interacting with the environment. An MDP $\mathcal{M}$ is defined as a tuple $(\mathcal{S}, \mathcal{A}, r, \mathcal{P}, \gamma, T) $, where $\mathcal{S}$ is a set of possible states, $\mathcal{A}$ is the set of actions, $\mathcal{P}$ is the transition probability distribution, $r$ is the reward function, $\gamma \in [0, 1]$ is the discount factor and T is the time horizon per episode. In RL, at each timestep $t$, the environment produces a state observation $s_{t} \in \mathcal{S}$. The agent then samples an action $a_t \sim \pi(s_t)$, $a_t \in \mathcal{A}$ and applies the action to the environment. As a consequence, the agent transitions to a new state $s_{t+1}$ sampled from the transition function $p(s_{t+1} | s_{t}, a_{t})$, $p \in \mathcal{P}$ or terminates the episode at state $s_{T}$. The agent's goal is to learn a stochastic behaviour policy $\pi$ parametrised by $\phi$, $\pi_\phi: \mathcal{S} \rightarrow \mathcal{P}(\mathcal{A})$ to maximise the expected future discounted reward $E[\sum_{i=0}^{T-1} \gamma^{i} r_i]$.

% In this work, we consider a state-of-the-art DRL algorithm, Proximal Policy Optimisation (PPO) \cite{schulman2017proximal}. 
% Further, we use GAIL to pre-train an agent with task demonstrations to extract the underlying motion patterns to effectively satisfy the RL objective.

\subsection{Proximal Policy Optimisation (PPO) \label{sec:PPO}}
% Policy gradient methods are a robust class of continuous control techniques that directly maximize the expected sum of rewards. The vanilla policy gradient estimator is given by 
% \begin{equation}
% L_{DRL} = -\nabla_{\phi} J_{PG} = - E_{\tau_\phi}[\sum_{t} \nabla_{\phi} \log \pi_{\phi}(a_t|s_t)A_t]
% \end{equation}
% where $\tau_\phi$ are the trajectories induced by the stochastic policy $\pi_{\phi}$ and $A_t$ is the advantage function. PPO is an approximation that reduces the variance of policy gradients by adaptively regulating the constraint on the Kullback-Leibler (KL) divergence of the old and new policies, which has shown robustness towards hyper-parameters \cite{schulman2017proximal}. PPO is an on-policy algorithm that learns the value of the policy being carried out by the agent including the exploration steps. It can be used for discrete action spaces.

PPO is an on-policy RL algorithm capable of dealing with both continuous and discrete action spaces. PPO alternates between collecting new observations and improving the policy, while approximating the value function as well \cite{schulman2017proximal}. The update function for the PPO policy is the following

\begin{footnotesize}
\begin{equation}
L(s_t,a_t,\theta_k, \theta)=min\left(\frac{\pi_{\theta}(a_t|s_t)}{\pi_{\theta_k}(a_t|s_t)} \hat{A}^{\pi_{\theta_k}}(s_t,a_t),g(\epsilon, \hat{A}^{\pi_{\theta_k}}(s_t,a_t))\right)
\end{equation}
\end{footnotesize}
where $\theta_k$ are the parameters of the old policy, and $g$ is defined as:

\begin{footnotesize}
\begin{equation}
g(\epsilon,\hat{A}) = \begin{cases}
    (1+\epsilon)\hat{A}_t, & \hat{A}_t\geq 0\\
    (1-\epsilon)\hat{A}_t,              & \hat{A}_t<0
\end{cases}
\end{equation}
\end{footnotesize}

Where $\hat{A}_t$ is the advantage estimator function at timestep $t$ and $\epsilon$ is a hyperparameter. The idea behind PPO is to limit the impact of the policy update using the min operator, so that the improvement in the policy is stable.
\subsection{Generative Adversarial Imitation Learning (GAIL)\label{sec:gail}}
GAIL is an imitation learning algorithm \cite{ho2016generative}. The principle of GAIL is based on the generative adversarial networks, which consists of a discriminator $D_{\phi'}$ and a policy generator $G_{\phi}$, where $\phi$ denotes the parameters associated with each network. The policy network generates exploration trajectories which are used by the discriminator to compute a surrogate function measuring the similarity between the generated policy and the expert policy. This similarity metric acts as reward proxy for the RL step. Unlike IRL techniques, GAIL directly generates policies instead of the reward function. The discriminator is trained to minimize the loss function:
\begin{equation}
L_{GAIL} = E_{\tau_{\phi}}[\log(D_{\phi'}(s_{t}, a))] + E_{\tau_{E}}[\log(1-D_{\phi'}(s_{t},a))]
\end{equation}
where $\tau_\phi$ are the trajectories generated by $G_\phi$ and $\tau_{E}$ are the expert trajectories. The policy generator $G_{\phi}$ is often adopted from methods based on stochastic policy such as PPO. There are two reasons why PPO is used for GAIL: first, PPO uses smooth policy update for stable learning and second, PPO generates diversified trajectories that act as a wide sampling range for the discriminator in GAIL.
% first, its smooth policy update for learning stability and second,  the diversity of the policy to sample in a wide range for the discriminator in GAIL.
\section{METHODS\label{sec:methods}}
The objective of our task is to efficiently and optimally train the robotic agent to expose the tumour region by grasping and retracting the fat tissue surrounding it. In particular, our DRL agent is represented by the End-Effector (EE) of the da Vinci Patient Side Manipulator (PSM) arm, which learns to move from an initial position $\mathbf{p}_0$ to a position close to the tumour $\mathbf{q}$, grasp the tissue and retract it to reach the desired target position $\mathbf{p}_T$. For the sake of simplicity, the PSM orientation is kept constant. The initial state of the anatomical environment, as well as the tumour centroid position $\mathbf{q}$, are assumed to be known from pre-operative data. 
% For the sake of simplicity, we have considered the orientation of the end-effector as constant.
In order to make the learnt motion primitives robust to different initial configurations, the EE starts from a different position after each episode (i.e. 2500 timesteps in our case) at training time. 
The considered state-space leverages solely on kinematics information defining the current robot state and the environment: 
\begin{equation}
S_t = [\mathbf{p}_t, \mathbf{q}, \mathbf{p}_T, ||\mathbf{p}_t-\mathbf{q}||, ||\mathbf{p}_t-\mathbf{p}_T||, g_t] \\
\end{equation}
where $|| . ||$ represents the Euclidean distance, $p_t$ denotes the position of the EE at time $t$ and $g_t \in \{0,1\}$ represents the gripper state (open/closed). 
At each time, the action the agent can take is defined by an increment motion of $0.5\beta\,mm$ in each spatial dimension, where $\beta$ varies in $\{0,-1,+1\}$, corresponding respectively to the conditions of no motion, backward motion and forward motion.

The presented algorithms are trained within \textit{UnityFlexML}\footnote{Project- https://gitlab.com/altairLab/unityflexml} framework, a platform to create simulation environments supporting deformable objects and dVRK kinematics. 

\subsection{Deep Reinforcement Learning (DRL) setup}
Standard DRL algorithms are trained in a simulation environment that is an exact replica of the real one.
% This simulation environment is promising for training an agent in simulation and deploying the learnt behaviour to the real system, considering simplified settings with a fixed starting position. 
We consider PPO as the standard baseline DRL algorithm \cite{schulman2017proximal}. For the training phase, we rely on a reward function which varies with the gripper state, to encourage EE motion towards the tumour if the tissue has not been grasped yet (i.e. gripper is still open), and tissue retraction if the tissue has been already grasped:
$$
r(s_t)=
	\begin{dcases*}
	- ||\mathbf{p}_t-\mathbf{q}||* k - 0.5, & \text{before grasp}\\
	- ||\mathbf{p}_t-\mathbf{p}_T||* k, & \text{after grasp}
	\end{dcases*} 
\eqno{(2)}
$$
The scalar quantity of $-0.5$ is added to restrict the reward in the range $(-1.0,-0.5)$ before grasping and $(-0.5,0)$ after grasping. The normalization constant $k$ is introduced to allow re-scaling of the trajectories to different working spaces and is inversely proportional to the maximum distance the PSM can move.

\subsection{Learning from Demonstrations (LfD) setup}
Since we aim at finding a better policy than the provided demonstrations, the training is based on the linear combination of the respective DRL and GAIL losses:
\begin{equation}
L_{Total} = \alpha L_{DRL} + \beta L_{GAIL} \\
\end{equation}
where $\alpha$ and $\beta$ represent weighting factors for the two loss functions. For a PPO agent, $\alpha=1$ and $\beta = 0$. For the GAIL agent, our initial investigation on hyper-parameters tuning yielded best performance for $\alpha=0.2$ and $\beta = 0.8$. Other values of $\alpha$ and $\beta$ yielded slower convergence.

Training of a GAIL agent requires the collection of trajectory demonstrations. In this work, task demonstrations are acquired on the real dVRK and transferred to the \textit{UnityFlexML} framework. The acquired trajectories consist of repetitive fat lifting task, performed by an expert user. Since the expert user is well aware of the final objective of tumour exposure, the grasp position is near the tumour for all the demonstrations. Moreover, the expert user is instructed to diversify the trajectories by starting each demonstration from a different initial position above the fat surface.

\begin{figure}[thpb]
	\centering
	\includegraphics[width=0.49\textwidth]{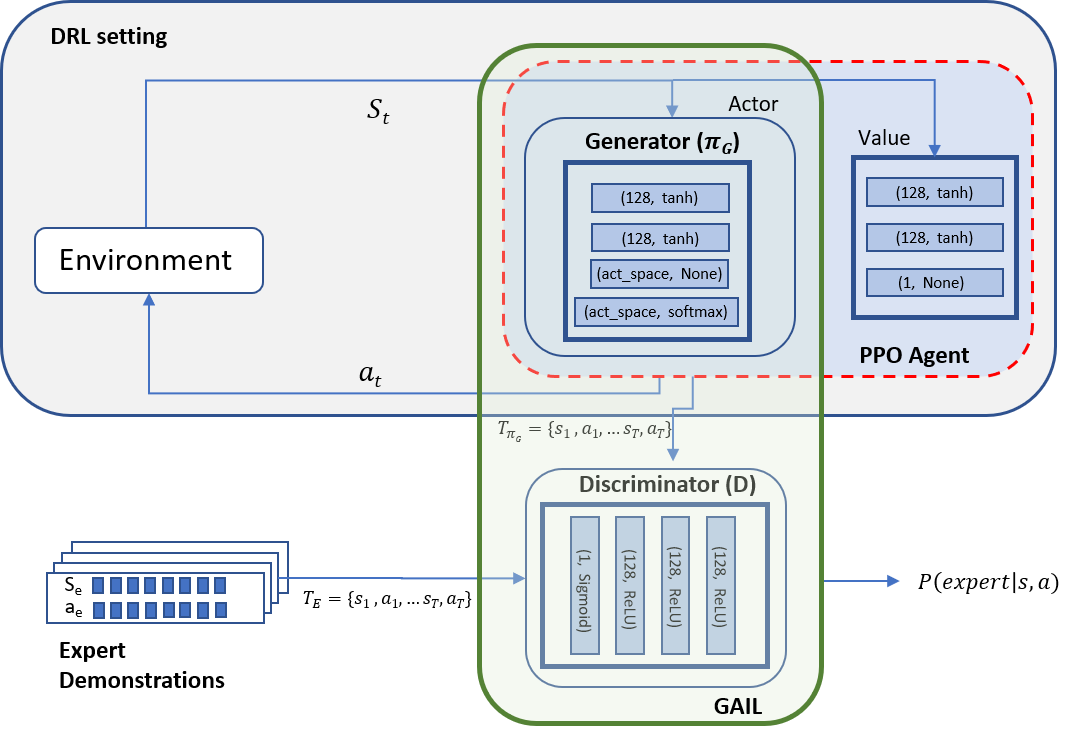}
	\caption{Network architecture of GAIL and PPO. PPO consists of a policy (actor) and Value network. The policy network acts as Generator for GAIL. Generated trajectories and expert trajectories are passed to the Discriminator. Discriminator learns a probability function which classifies the generator trajectory as expert or non-expert. The network layer details are depicted inside each box in the format (hidden units, activation) respectively.} 
	\label{fig:gailarc}
	%%%\vspace{-2 mm}
\end{figure}

% Acquisition of task demonstrations from the real environment leverages on a connection between \textit{UnityFlexML} and ROS, using the established method described in \cite{qian2019dvrkxr}.
Acquisition of task demonstrations from the real environment leverages the communication pipeline provided by \textit{UnityFlexML} (Fig.~\ref{fig:setup}). Registration between the simulated and real environment is guaranteed following the same registration process described in \cite{tagliabue2020soft}.
The joint values are sent to \textit{UnityFlexML} through UDP sockets and the desired configuration is reached with direct kinematics. Each recorded demonstration consists of the set of kinematic observations that define the state space (Sec.~\ref{sec:methods}) and the corresponding actions at each timestep. An important aspect of this implementation is the challenge associated when we reset each episode. 
% In the simulation, the position of the EE is teleported to the initial point of the next episode whenever the episode resets. However, in a real robotic system it is unfeasible to do the same.
% Hence, while recording demonstrations, we change the reset condition when starting a new episode. In the new scenario, the agent resets with a delay of some timesteps after the grasp is released by the human operator. This gives us enough time to release the grasp and move to a new random point to start the new episode. 
In the simulation, as soon as the target position is reached, the grasp is released and the episode resets.  The position of the EE is then immediately teleported to the next initial point. 
This reset strategy has been adapted to cope with the real robotic system, during the recording of expert demonstrations. In particular, a delay of some timesteps has been added between the moment when the grasp is released and the beginning of the next episode, to allow repositioning.
We make use of 35 continuous episodes recordings. 
Although our simulation framework supports demonstration recordings using a keyboard or a joystick, the established communication pipeline between dVRK and \textit{UnityFlexML} is crucial since it helps to acquire demonstrations directly with the real robotic system, thus without deviating from the surgical workflow.
The network architecture used for our proposed GAIL and PPO training method is depicted in Fig.~\ref{fig:gailarc}.

\begin{figure*}[thpb]
	\centering
	\includegraphics[width=0.8\textwidth]{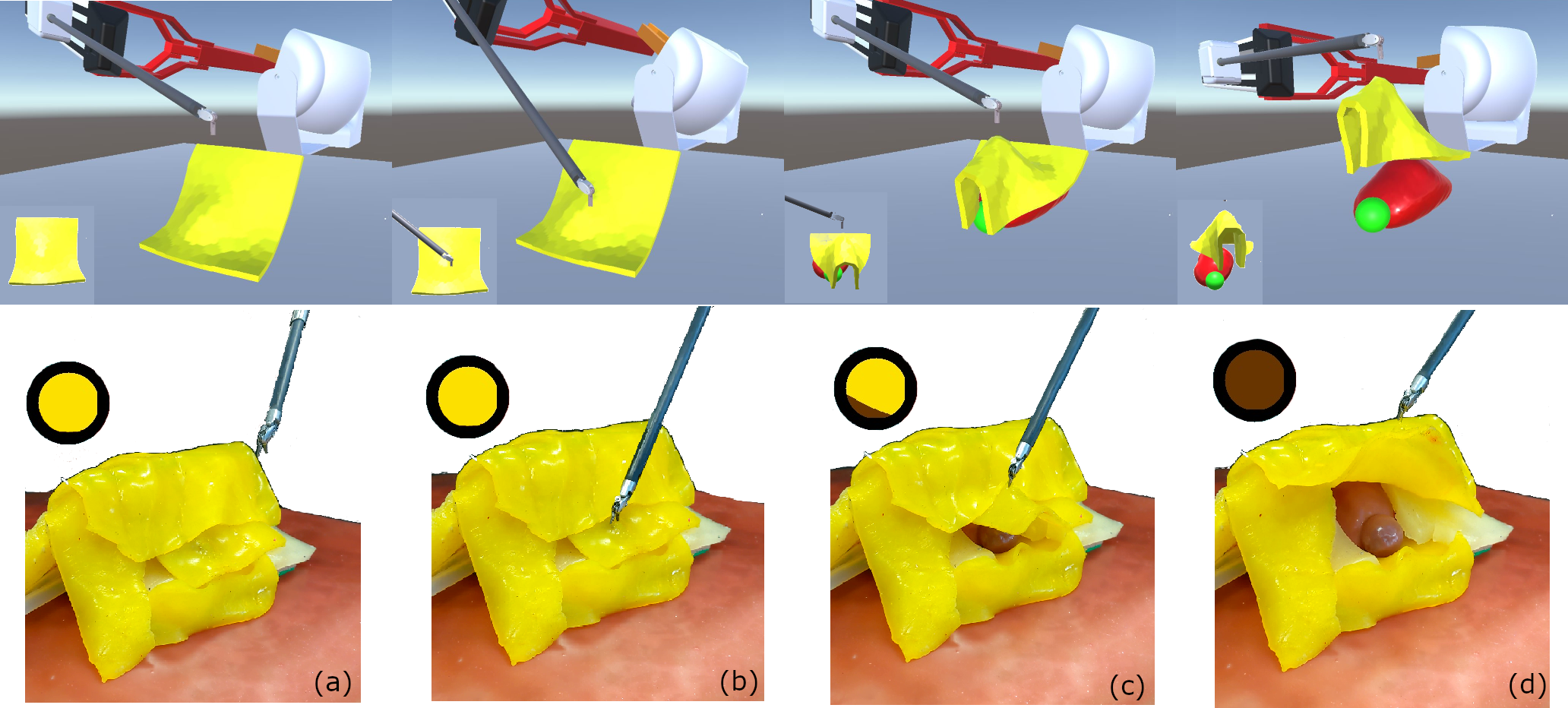}
	%%%\vspace{-2 mm}
	\caption{Sequence of action frames for task completion in simulation (top) and reality (bottom). From left to right: approach, grasp, retract, tumour exposure. (top) Perspective of the simulated camera is overlaid on the bottom left of the simulator frames. (bottom) The real camera is placed in the same position as in the simulation (which do not correspond to the viewpoint used to take these pictures)
	The colour segmentation using a circular mask is illustrated for all the configurations (a) 0\% exposure (b) 0\% exposure (c)$\sim$15\% exposure (b) 100\% exposure.} 
	\label{fig:simpixels}
	%%%\vspace{-1 mm}
\end{figure*}
% %%%\vspace{-1 mm}
\section{EXPERIMENTS\label{sec:experiments}}

We consider a tissue retraction task in the context of a partial nephrectomy procedure. In particular, we aim at exposing a kidney tumour that is hidden by perinephric fat tissue. The real robotic setup consists of a synthetic kidney phantom covered with silicone fat tissue (Fig.~\ref{fig:simpixels}). Methods presented in Sec.~\ref{sec:methods} are trained in a simulation environment which is an exact replica of the phantom, where the deformation properties of the fat tissue have been optimized to mimic those of the real synthetic tissue used in the experiments \cite{tagliabue2020soft}. All the simulation experiments, including DRL training and dVRK control, are executed on a  workstation equipped with an AMD Ryzen 3700X processor and  NVIDIA TitanX GPU.

The considered algorithms are tested both in simulation and reality based on two different criteria: sample efficiency and the optimality of the accomplished task, i.e. the ability of each method to expose the tumour. Sample efficiency is estimated by the number of time steps required by each algorithm to reach high reward values. Secondly, we estimate the optimality of the learnt behaviour by a Tumour Exposure (TE) metric. TE is assessed as the normalized percentage of tumour surface which can be seen from a camera positioned in front of the kidney, both for the simulated and the real setup. This evaluation allows us to understand if the quality of the exposure can be maximized regardless of the PSM starting position.
% The camera point of view can be appreciated in Fig.3 top (simulated) and Fig 6 (real).
The behaviour learnt in the simulated and real scenario is depicted in Fig.~\ref{fig:simpixels} top and Fig.~\ref{fig:simpixels} bottom, respectively.

\subsection{Simulation experiments \label{sec:experiment_Sim}}
The high level of realism of the simulated environment created within \textit{UnityFlexML} does not only allow us to train the standard DRL methods with a sim-to-real approach but also provides a platform for testing the presented methods in realistic settings. 
In order to assess whether the behaviour learnt by the agent is robust to different starting EE positions, we perform an experiment where the trained agent has to perform the task starting from 49 different positions uniformly sampled on a 7x7 regular grid above the portion of the fat tissue. We evaluate TE each time the EE reaches $\mathbf{p}_T$.
  
\subsection{Real robotic experiments \label{sec:experiment_real}}
Our real experimental setup (Fig.~\ref{fig:simpixels} bottom) is precisely aligned with the simulated scene (Fig.~\ref{fig:simpixels} top) by mapping the poses of the PSM in the two environments, as described in \cite{piccinelli2020}. Accurate registration is essential to prevent inconsistencies between the two environments, especially considering that all the movements of the dVRK arm in the real system are controlled through the simulated robot, including the grasping action.
To compute the TE metric, we select a circular region of interest around the tumour (exploiting the fact that its position is fixed) and we extract the visible portion by applying a mask with HSV bounds matching tumour colour (Fig.~\ref{fig:simpixels} bottom).

\section{RESULTS AND DISCUSSION\label{sec:results}}
The results obtained will be presented in two separate sections, the first dedicated to simulation results while the second is dedicated to the results on the real setup.

\subsection{Simulation \label{sec:results_sim}}
\begin{figure}[thpb]
	\centering
	%%%\vspace{-5 mm}
	\includegraphics[width=0.49\textwidth]{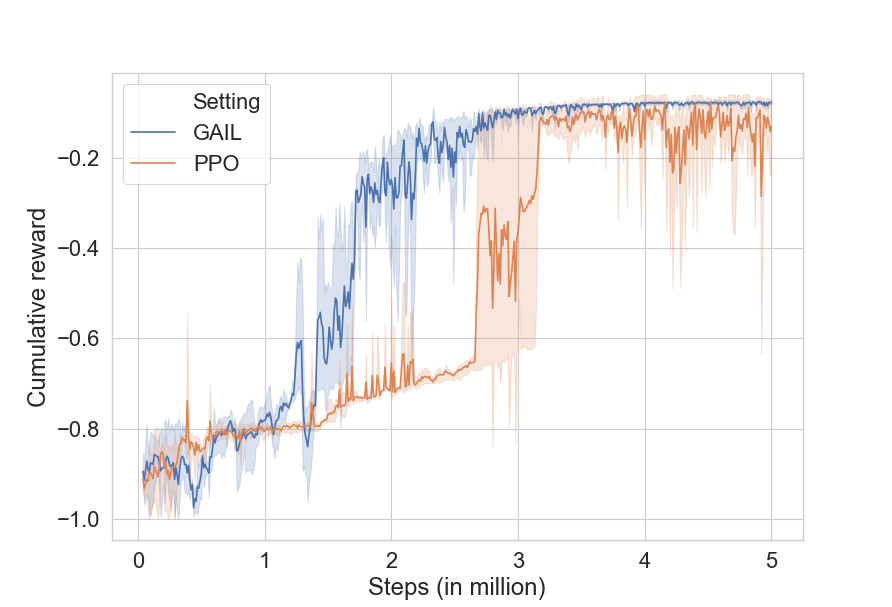}
	%%%\vspace{-5 mm}
	\caption{The obtained learning curve for GAIL and PPO. Cumulative reward is normalised in the range $[-1,0]$. The shaded area spans the range of values obtained when training the agent starting from three different initialization seeds.} 
	\label{fig:curve1}
	%%%\vspace{-3 mm}
\end{figure}

\begin{figure}[thpb]
	\centering
	\includegraphics[width=0.49\textwidth]{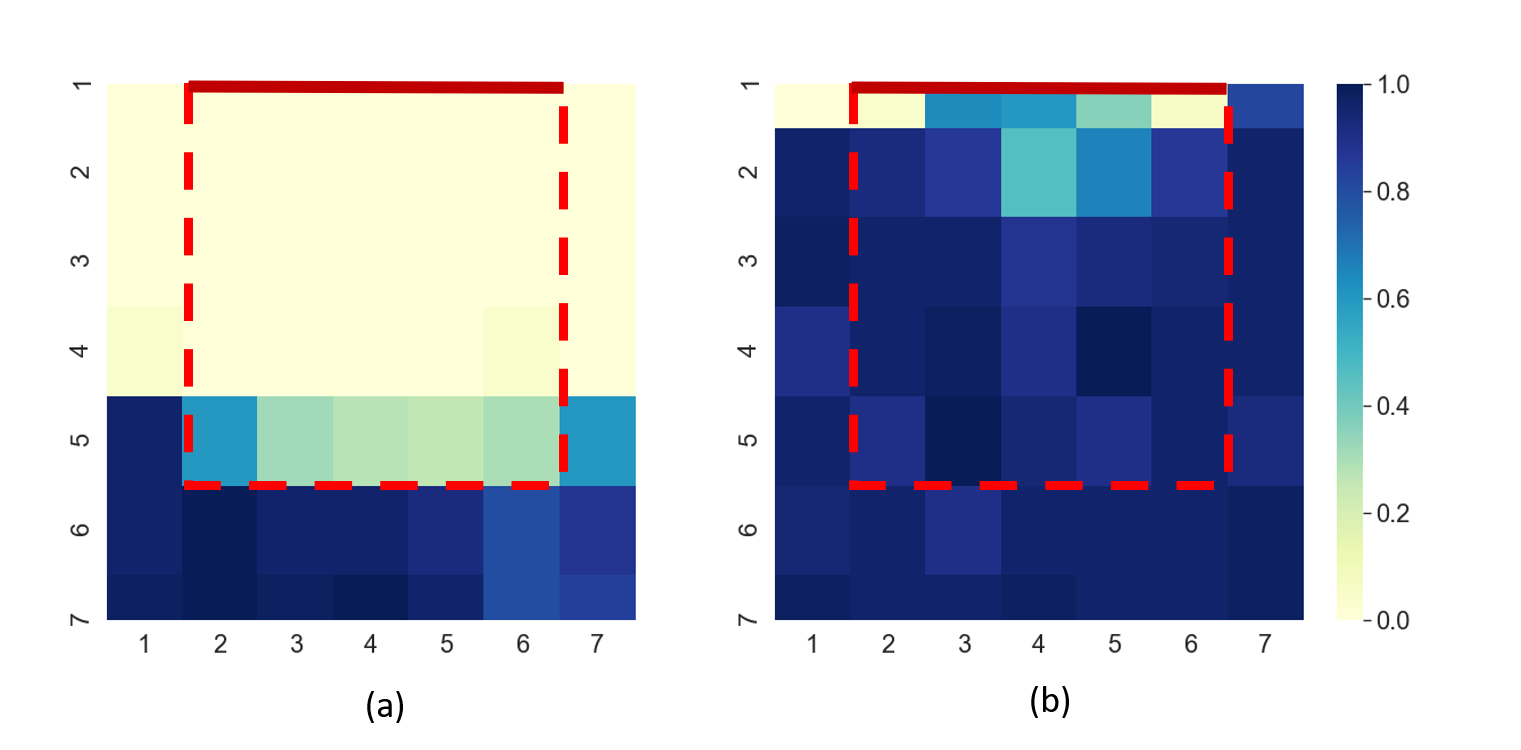}
% 	%%%\vspace{-5 mm}
	\caption{Simulation experiments: TE from the camera at different initial positions, of the PSM for (a) PPO, (b) GAIL. The colour of each subregion is related to the percentage of visible tumour area when $\textbf{p}_0$ belongs to that subregion. The fat boundary from the top view is depicted in red dashed lines whereas the fat attachment is shown in the solid red line} 
	\label{fig:heat_simulation}
	%%%\vspace{-0 mm}
\end{figure}
 
% Fig \ref{fig:curve1} shows the learning curves obtained with the considered learning configurations. 
Learning curves obtained with the considered learning configurations are showed in Fig.~\ref{fig:curve1}.
GAIL is more efficient than PPO and shows a monotonous and smooth learning pattern. 
% It starts learning (moves towards high-reward values) and diverges from baseline PPO around 1 million steps. 
GAIL learning curve begins to increase towards high-reward values and diverges from PPO around 1 million steps. PPO shows a modular reward trend: it requires 2.5 million steps to learn the approach behaviour and 1 million steps to learn the retract behaviour.
This experiment shows that incorporating human demonstrations makes learning sample-efficient compared to baseline PPO. This result verifies our hypothesis that incorporating human knowledge can provide initial prior reference and takes fewer timesteps to learn the behaviour.

\begin{figure}[thpb]
	\centering
	\includegraphics[width=0.45\textwidth]{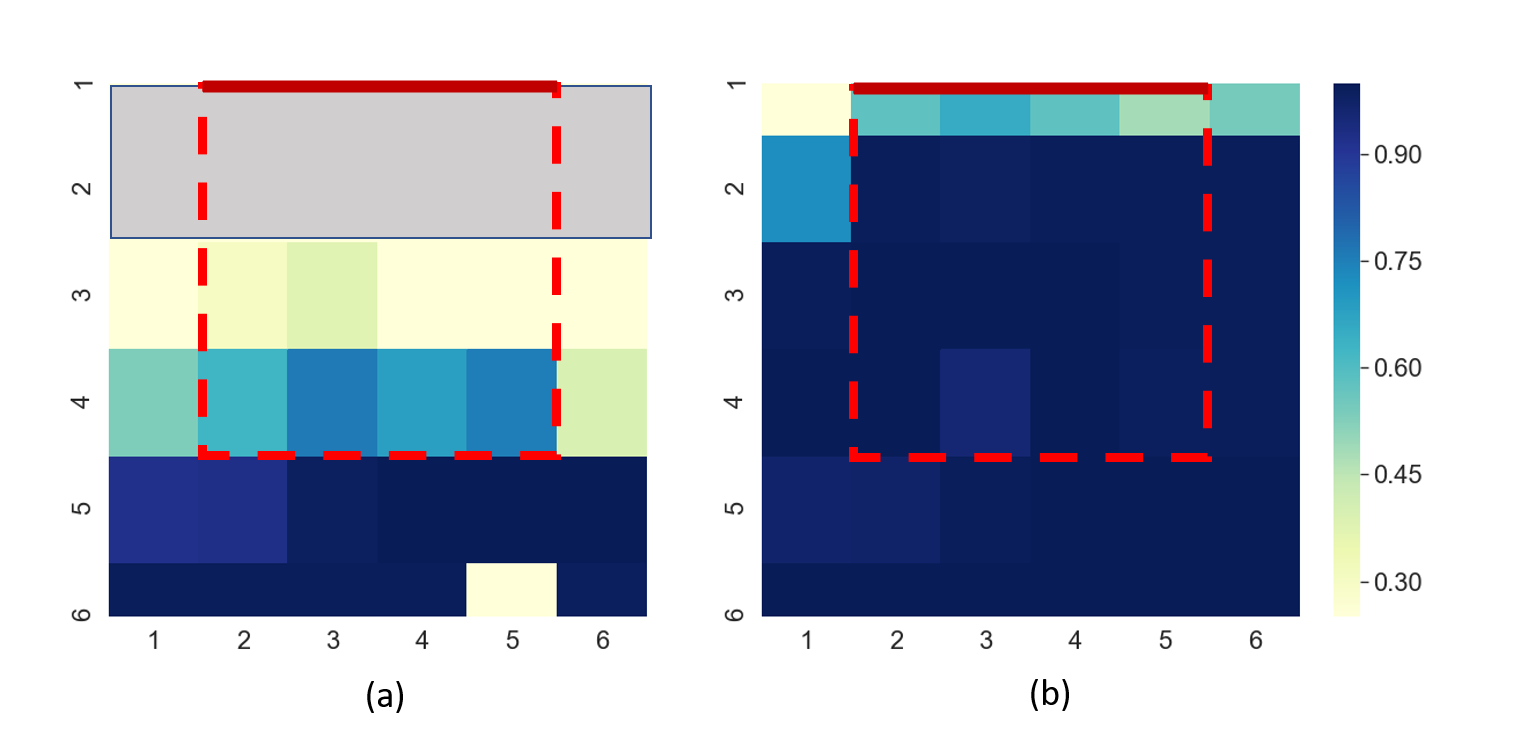}
	%%%\vspace{-2 mm}
	\caption{Real grasp experiments: TE from the camera when starting from different initial positions of the EE, using (a) PPO (b) GAIL. The portion of fat tissue which is not considered for the experiments is coloured in grey.} 
	\label{fig:heat_real}
	%%%\vspace{-2 mm}
\end{figure}

The plot in Fig.~\ref{fig:heat_simulation} shows the TE from the simulated camera depending on the starting position of the PSM arm above and outside the boundary of the fat tissue and points. In the case of PPO, when the starting EE position is close to the fat attachment, the agent tends to grasp near the proximity of the fat attachment, thus causing little or no tumour exposure (Fig.~\ref{fig:heat_simulation}a). Note that, for PPO, the reward function is handcrafted such that before the grasp, the agent is encouraged to approach the known position of the tumour, because grasping near the tumour is the best strategy to maximise tumour exposure. 
% Further, the function changes drastically to a higher value upon grasping fat to initiate the retract behaviour. The learnt motion of the agent and the reward curve indicates that the algorithm learns to maximise the cumulative reward while showing less TE. 
Low TE indicates that manually tuning the reward function to encode complex task objective such as tumour exposure can be challenging. We performed a preliminary evaluation of a scenario where we design a reward function that incorporates a TE-dependent term. However, it did not show significant improvements in the results.
% Our initial evaluation to introduce the TE factor in the reward function showed no significant improvement compared to the reward without the TE exposure parameter. 
This might be due to the fact that before grasping the tumour exposure is always zero and represents a sparse reward scenario. We plan to further investigate this in future works. 

On the contrary, when human demonstrations are incorporated, GAIL is able to grasp closer to the tumour and expose the tumour regardless of the starting position. Note that the strategy adopted by the user while acquiring demonstrations is to move and grasp towards points close to the tumour to maximize the exposure.
The difference between the behaviour learnt by PPO and GAIL is in the robust selection of the grasping point when the starting position varies. When the starting position is above the attached area, PPO tends to grasp near the fat attachment, thus leading to low TE, whereas GAIL learns to grasp closer to the tumour, obtaining a higher TE.
% The difference between the trajectories learnt by PPO and GAIL is the grasping point. PPO grasps near the fat attachment which leads to low TE, whereas GAIL grasps near the tumour leading to high TE.
% since it learns to imitate the human operator, who moves towards points close to the tumour to maximize the exposure (Fig \ref{fig:heat_simulation}b). 
% This demonstrates that using demonstrations, GAIL learns the grasping point in a robust manner starting from various initial positions.
% trajectory features to perform human-like generalised motion from various initial points. 

\subsection{Real robotic setup \label{sec:results_real}}
We have been able to successfully replicate the learned behaviour from the simulated to the real environment without any appreciable inconsistency. The da Vinci EE successfully gets in contact with the fat tissue for all the different initial positions, and it is always able to reach the target point. The tumour exposure percentage starting from various points is illustrated in Fig.~\ref{fig:heat_real}. 
For PPO, we did not initialize the EE positions near the attachment (represented as the unattempted grey region in Fig.~\ref{fig:heat_real}a) because grasping near the attachment led to fat tissue tear. 
When comparing the results obtained for GAIL and PPO, it emerges that GAIL is not only able to reach higher overall exposure, but it is also more robust to changes in the initial PSM position. In particular, tumour exposure is achieved also when starting from points that were unattempted for PPO (Fig.~\ref{fig:heat_real}b), thus suggesting an overall improvement in performances, due to grasping closer to the tumour position.

This observation indicates that the initial PSM position has a great impact on the performances in the case of PPO, whereas GAIL is able to reach optimal performance regardless of the starting position, confirming results obtained in the simulated experiments.
In terms of the Average TE (ATE) computed considering all trials from different starting points, PPO obtains an ATE of 0.33 in a simulated environment and 0.38 in the real, while GAIL obtains an ATE of 0.84 in simulation and 0.90 in the real domain. We can infer that performance using demonstrations in both simulation and real-world is robust and outperforms PPO.

\section{CONCLUSIONS \& FUTURE WORK\label{sec:conclusion}}

In this work, we present an LfD methodology for automation of tissue retraction based on GAIL. The proposed methodology can be trained in a simulated replica of the real scene using a small set of real robotic demonstrations and deployed in the real environment. The method builds on a consolidated DRL architecture and can learn generalised human-like trajectories in a sample-efficient way. Experiments in simulation and real environment show that, while both baseline DRL methods and GAIL can accomplish the task, the latter boosts the learning process by reducing the number of steps required and learning near-human trajectories. The learnt policy has been deployed on the dVRK and the tissue retraction task has been successfully completed.

This work has some limitations. The underlying hypothesis is based on the assumption of knowing the target positional coordinates (e.g., tumour position) pre-operatively. However, the tissue retraction surgical gesture can be carried out as an exploratory subtask without a known target. Hence, our future work will be directed in utilizing visual information to estimate the kinematic coordinates of the various image features, as described in \cite{attanasio2020autonomous, li2020super}. Further experimentation will be carried out to assess the impact of the quality and the number of demonstrations required to learn optimal behaviour with different surgical gestures, involving experts from various levels of expertise. Another limitation to consider is the safety issues due to free exploration of the state space that might lead to dangerous movements. Hence, in subsequent work, we will incorporate safety constraints through a Safe-Reinforcement learning technique \cite{cheng2019end}.

In conclusion, surgical subtask automation is an anticipated research direction that will lead to a higher level of autonomy and put the surgeons in a supervisory role. This approach will enable a reduction in surgical workload and improved patient outcomes. 

\section*{ACKNOWLEDGEMENT}

The authors would like to thank Enrico Magnabosco, Sanat Ramesh for their support in the initial development of the simulator and discussion respectively. We also acknowledge the support of NVIDIA Corporation for the donation of the Titan Xp GPU used in this research.
\bibliographystyle{IEEEtran}
\bibliography{root}
\end{document}